\documentclass[peerreview, draftcls, onecolumn]{IEEEtran}

\usepackage{cite, graphicx, amsmath, amsfonts, amssymb, array, latexsym, url, algorithm, algorithmic}
\usepackage{multirow, dsfont}

\newcommand{\xx}{{\bf x}}
\newcommand{\Aa}{{\bf A}}
\newcommand{\Hh}{{\bf H}}
\newcommand{\argmax}{\operatornamewithlimits{argmax}}
\newcommand{\argmin}{\operatornamewithlimits{argmin}}

\begin{document}

\title{Iterative Shrinkage Approach to Restoration of Optical Imagery}

\author{Elad~Shaked,~\IEEEmembership{Student Member,~IEEE} and Oleg~Michailovich,~\IEEEmembership{Member,~IEEE}%
\thanks{Manuscript received Month XX, 2009. This research was supported by a Discovery grant from NSERC -- The Natural Sciences and Engineering Research Council of Canada. Information on various NSERC activities and programs can be obtained from {\tt http://www.nserc.ca}.}%
\thanks{E. Shaked and O. Michailovich are with the School of Electrical and Computer Engineering, University of Waterloo, Canada N2L 3G1 (phone: 519-888-4567; e-mails: {eshaked, olegm}@uwaterloo.ca).}}

\maketitle

\begin{abstract}
The problem of reconstruction of digital images from their degraded measurements is regarded as a problem of central importance in various fields of engineering and imaging sciences. In such cases, the degradation is typically caused by the resolution limitations of an imaging device in use and/or by the destructive influence of measurement noise. Specifically, when the noise obeys a Poisson probability law, standard approaches to the problem of image reconstruction are based on using fixed-point algorithms which follow the methodology first proposed by Richardson and Lucy. The practice of using these methods, however, shows that their convergence properties tend to deteriorate at relatively high noise levels. Accordingly, in the present paper, a novel method for de-noising and/or de-blurring of digital images corrupted by Poisson noise is introduced. The proposed method is derived under the assumption that the image of interest can be sparsely represented in the domain of a linear transform. Consequently, a shrinkage-based iterative procedure is proposed, which guarantees the solution to converge to the global maximizer of an associated  maximum-a-posteriori criterion. It is shown in a series of computer-simulated experiments that the proposed method outperforms a number of existing alternatives in terms of stability, precision, and computational efficiency.
\end{abstract}


\IEEEpeerreviewmaketitle

\section{Introduction}\label{Intro}
The image formation model of many key imaging modalities rely on the notion of {\em event counts}. The latter, for example, quantifies the number of gamma photons which pass though a single slit of the collimator of a gamma camera in positron emission tomography (PET) and single positron emission computer tomography (SPECT)~\cite{Lee95, Yavuz98, Bauschke99}. In addition, statistical models of the same type are routinely used in optics to account for the process of ``counting" the number of optical photons registered by a sensor of a (CCD) camera~\cite{Boie92, Healey94}. Confocal microscopy~\cite{Holmes92}, astronomical imaging~\cite{Bradt04}, and turbulent imaging~\cite{Roggeman96} are among other fields of imaging sciences where the notion of event counts is standardly used.

In the most general setup, the image formation model of the above-mentioned imaging modalities assumes that a data image is formed as a blurred version of the original (i.e. true) image contaminated by Poisson noise. Specifically, if $g(\xx) \in \mathbb{Z}_{+}$ (with $\xx \in \Omega \subseteq \mathbb{Z}^2$) denotes the observed Poisson counts image and $f(\xx) \in \mathbb{R}_{+}$ is its corresponding original counterpart, then the image formation model can be formally expressed as given by~\cite[Section 7.3]{Bertero98}
\begin{equation}\label{E1}
g = \mathcal{P} \left\{ {\bf H}[f] \right\},
\end{equation}
where $\mathcal{P}$ stands for the operation of contamination of ${\bf H}[f]$ by Poisson noise and $\bf H$ denotes the operator of convolution with a (known) point spread function (PSF) $h$, where $h$ is assumed to be positive and mean preserving. Accordingly, to recover the original image $f$ from $g$, the combined effect of $\mathcal{P}\{{\bf H} [\cdot]\}$ has to be inverted.

The model of (\ref{E1}) has long been in use in a variety of different applications. For example, in \cite{Holmes92} the model describes the formation of microscopic images, which are subsequently enhanced by means of a maximum-likelihood (ML) procedure, also known as the Richardson-Lucy algorithm \cite{Richardson72, Lucy74}. The same model and the same reconstruction approach has been used in the field of nuclear imaging, where they have been employed for recovery of PET and SPECT scans \cite{Shepp82, Mignotte02}. It was argued in \cite{Dey06}, however, that the ML approach may lead to either unstable or noisy results in the case of poorly conditioned operators $\bf H$. To alleviate this problem, it was proposed in \cite{Dey06} to replace the ML approach by maximum-a-posteriori (MAP) estimation, which provides means to regularize the inversion of $\bf H$ via incorporating some prior information about $f$ in the process of its reconstruction. Specifically, \cite{Dey06} suggested using either the Tikhonov-Miller or bounded variation models for $f$. Unfortunately, as will be shown in the experimental part of the present paper, the above algorithms cannot fully alleviate the instability problem intrinsic in the inversion of (\ref{E1}). To overcome this deficiency, the work in \cite{Figueiredo09} suggested a different approach also based on the bounded variation model which leads to a minimum {\em total-variation} (TV) solution for the true image. In this case, to find a solution to (\ref{E1}), a variable splitting procedure of \cite{Goldstein09} was employed. Despite a substantial improvement in the stability of reconstruction, the computational cost of \cite{Figueiredo09} is relatively high, which undermines the applicability of this method for processing large data sets. Moreover, the algorithm of \cite{Figueiredo09} may produce non-positives solutions, which are not natural in the case of Poisson imaging. To overcome the computational inefficiency of \cite{Figueiredo09}, a modification of the variable splitting method was proposed in \cite{Setzer09}. Unfortunately, this modification seems to ``restore" the instability concerns.

A different class of reconstruction methods has been recently proposed based on the assumption of {\em compressibility} of the true image $f$. This assumption suggests that $f$ can be {\em sparsely represented} either in the spatial domain or in the domain of a certain linear transform. Thus, for example,  \cite{Lingenfelter09} analyzes three different penalty functions, whose role is to impose the sparsity constraint on the estimated images. Although the proposed algorithms have exceptional stability and computation properties, they require the original image to be sparse in the spatial domain -- the assumption which does not seem to be natural for a large spectrum of practical images. In this regard, a more general assumption has been used in \cite{Dupe08}, where the true images are supposed to be sparsely represented in the domain of a tight frame. This method, however, is based on ``gaussianization" of the Poisson noise by means of a {\em variance stabilization transform} (VST) \cite{Anscombe48} that is known to perform inadequately in low-count settings. Another work that employed the assumption on $f$ to be sparsely representable in the domain of an orthonormal transform is \cite{Harmany09}, where the sparsity-constraned Poisson inverse problem is solved in an approximate way by minimizing a sequence of $\ell_2 - \ell_1$ objective functions. However, as acknowledged by the authors of \cite{Harmany09}, their algorithm has a drawback of slow convergence, which should be addressed in future research. As well, one should also mention the method of \cite{TheWitch03}, in which an estimate of the true image $f$ is obtained based on maximization of an ML criterion penalized by the $\ell_0$-norm of the platelet coefficients of the estimate. It is not clear, however, if the method of \cite{TheWitch03} could be used for a different type of image representations (rather than platelets), let alone the necessity to augment the performance of \cite{TheWitch03} by the computationally expensive cycle-spinning procedure of \cite{Antoniadis95} to optimize the quality of resulting estimates.

In the case when the effect of ${\bf H}$ can be neglected, a different family of reconstruction methods has been suggested based on the use of VST \cite{Anscombe48}. In particular, in such methods, the data image $g$ is first preprocessed by a VST that transforms the Poisson noise into approximately homogeneous and Gaussian \cite{Fryzlewicz04, Jansen06, Fryzlewicz07, Zhang08}. Subsequently, the true image is recovered from the ``gaussianized" data using an appropriate statistical framework. Among other reconstruction techniques for solving (\ref{E1}) with ${\bf H} \approx {\bf I}$ are wavelet Wiener filtering \cite{Nowak99}, hypothesis testing \cite{Kolaczyk00}, as well as empirical Bayesian approaches \cite{Timmermann99, Nowak00}.

In the current paper, a different method for reconstruction of the true image $f$ in (\ref{E1}) is introduced. As opposed to the algorithms described above, the proposed method is exceptional for it concurrently fulfills a number of essential objectives, {\it viz.}
\begin{enumerate}
\item{\it Exactness:} No auxiliary transformations and/or approximations are applied to the data image $g$ to modify the properties of measurement noise, and therefore the method is applied under realistic statistical assumptions regarding the noise nature.
\item{\it Generality:} The image formation model employed by the proposed method is designed to accommodate a number of reconstruction scenarios, namely de-noising, de-blurring, and a combination thereof. As a result, the very same procedure can be applied to recover an image of interest for a spectrum  of different degradations.
\item{\it Versatility:} The proposed reconstruction is carried out under the assumption that the original image  $f$ can be sparsely represented in the domain of a linear transform. This assumption is currently recognized to be superior to many alternative models, including the Tikhonov-Miller and total variation models~\cite{Elad07}. 
\item{\it Efficiency:} The proposed solution is based on the concept of {\em iterative shrinkage}~\cite{Daubechies03, Elad07, Blumensath08} as a modern, computationally efficient, and stable procedure for solving a variety of inverse problems. In this way, the present contribution extends the theory of iterative shrinkage (aka thresholding) to the case of problems concerned with Poisson noises.
\item{\it Uniqueness:} Under a certain setup, the proposed algorithm is guaranteed to converge to the global maximizer of a maximum-a-posteriori criterion, thereby providing a unique solution to the reconstruction problem at hand. 
\end{enumerate}

Moreover, in the experimental part of the paper, it is shown that the proposed reconstruction method outperforms a number of alternative algorithms in terms of normalized mean-square error (NMSE), SSIM quality index~\cite{Wang04}, as well as its stability and computational efficiency. 

The rest of the paper is organized as follows. Section II describes model assumptions and derives an optimization framework to be used in the following sections. Section III details a bound optimization approach, which results in an iterative shrinkage procedure. Section IV provides a proof of convergence of the proposed algorithm, while Section V presents a number of {\it in silico} reconstruction results. Section VI finalizes the paper with a discussion and conclusions.

\section{Estimation Framework}\label{EST}
\subsection{Image modeling}
The proposed approach for the reconstruction of $f$ in (\ref{E1}) is based on the assumption that $f$ can be sparsely represented in the domain of a certain linear transform. Specifically, let $\{\varphi_k\}_{k\in\mathcal{I}}$ be a {\em frame} in the space of the original image $f$ \cite{Christensen08}. (Note that $\mathcal{I}$ above denotes a set of frame indices, whose definition depends on a specific choice of $\varphi_k$.) Using the frame, $f$ can be represented according to
\begin{equation}\label{E2}
f = {\bf \Phi} (c): f \mapsto \sum_{k\in\mathcal{I}} c_k \varphi_k,
\end{equation}
where $c \in \ell_2(\mathcal{I})$ denotes the representation coefficients of $f$ in ${\rm span}\{\varphi_k\}_{k\in\mathcal{I}}$, such that the number of {\em significant} $c_k$ is much smaller than $\#\mathcal{I}$, i.e. the total number of $\varphi_k$. It is worthwhile noting that the model of (\ref{E2}) is standard in the theory of sparse representation which has firmly reserved a leading position among the modern tools of signal and image processing. The value of sparse representations has been demonstrated in numerous fields and applications, which include signal modeling~\cite{Chen01}, compressed sensing~\cite{Donoho06, Candes06}, independent component analysis and blind source separation~\cite{Bofill01, Georgiev05}, inverse problems~\cite{Daubechies03, Figueiredo03}, signal and image de-noising~\cite{Donoho95, Aharon06}, morphological component analysis~\cite{Starck04} as well as its earlier version in~\cite{Michailovich02}, just to name a few.

When expressed as a function of the representation coefficients $c$, the image formation model (\ref{E1}) becomes
\begin{equation}\label{E3}
g = \mathcal{P} \left\{ {\bf H}[{\bf \Phi} (c)] \right\} = \mathcal{P} \left\{ {\bf A}[c] \right\},
\end{equation}
where ${\bf A} = {\bf H} \cdot {\bf \Phi}$ is a composition map from $\ell_2(\mathcal{I})$ to the signal space, which represents the combined effect of image synthesis and blur. In the discussion below, we will be mainly interested in the following three settings:
\begin{itemize}
\item {\it Setting 1:} The blur is negligible and hence ${\bf A} \equiv {\bf \Phi}$. In this case, the problem becomes that of {\em image de-noising}.  
\item {\it Setting 2:} The basis $\bf \Phi$ is the canonical (Dirac) basis (i.e. ${\bf \Phi} = {\bf I}$) and hence ${\bf A} \equiv {\bf H}$. In this case, the image $f(\xx)$ is identified with its representation coefficients $c$, and the problem becomes that of {\em sparse deconvolution}.
\item {\it Setting 3:} Neither $\bf H$ nor $\bf \Phi$ can be simplified/neglected. In this case, the estimation problem at hand becomes that of {\em sparse reconstruction}.
\end{itemize}
For the sake of generality, in what follows the specific nature of $\bf A$ will not be specified until the experimental part of the paper, where different reconstruction examples are presented. In particular, $\bf A$ is considered to be a linear map from the discrete domain of the representation coefficients $c \in \ell_2(\mathcal{I})$ to the space of blurred images. 

It is well known that in the case of a poorly conditioned operator $\bf A$, the problem of recovering $c \in \ell_2(\mathcal{I})$ from noisy measurements of ${\bf A}[c]$ can be highly unstable (in the sense that there will not be a continuous dependency between the data and an estimate of $c$). In order to overcome this deficiency, we employ the framework of MAP estimation which provides the most likely solution, given the observed data and a reasonable assumption regarding the statistical nature of the true coefficients $c$~\cite{Carlin96}. The details on this approach are presented next.

\subsection{MAP estimation}
Since most of the modern imaging devices work with finite data, it seems to be appropriate to restrict the subset $\Omega \subset \mathbb{Z}^2$ to be a finite lattice. Accordingly, the measurement $g$ and the original image $f$ are assumed to be of size $N\times M$ and $K\times L$, respectively. In this case, the range of $\bf A$ becomes a set of bounded $N\times M$ matrices. 

To compensate for the information loss inflicted by the poor conditioning of $\bf A$, the MAP estimator takes advantage of {\it a priori} knowledge regarding the quantity of interest, i.e. $c$. In the case at hand, such knowledge is specified in the form of a prior probability distribution $P(c)$, which leads to the following definition of the MAP estimator
\begin{equation}\label{E4}
c_{MAP} = \argmax_c P(c \,|\, g)= \argmax_c P(g \,|\, c) \, P(c),
\end{equation}
where $P(g \,|\, c)$ represents the likelihood function pertaining to the observation model (\ref{E1}). In the case of Poisson noises, the likelihood function of a single measurement $g(\xx_i)$ at pixel $\xx_i$ is given by
\begin{equation}\label{E5}
P(g(\xx_i)\,|\, c) = \frac{e^{-(\Aa[c])_i}(\Aa[c])_i^{g(\xx_i)}}{{g(\xx_i)}!},
\end{equation}
where $(\Aa[c])_i$ denotes the $\xx_i$-th coordinate of ${\bf A}[c]$ and $g(\xx_i) \in \mathbb{Z}_+$ is interpreted as the ``number of counts" (e.g. the number of gamma photons registered by a gamma camera in nuclear imaging). Assuming that the values of $g(\xx_i)$ are independent and identically distributed (i.i.d.), the joint probability of the observed image $g$ is given by
\begin{equation}\label{E6}
P(g \,|\, c) = \prod_{\xx_i = 1}^{N M} \frac{e^{-(\Aa[c])_i}(\Aa[c])_i^{g(\xx_i)}}{g(\xx_i)!}
\end{equation}

To complete the model, the coefficients of $c=\{c_k\}_{k\in \mathcal{I}}$ are assumed to be identically distributed according to a Generalized Gaussian (GG) probability law~\cite{Moulin99}, so that the joint probability of $c$ can be defined as 
\begin{equation}\label{E7}
P(c) = \prod_{k \in \mathcal{I}}  \frac{p}{2 \, \beta \, \Gamma(p^{-1})} e^{-(|c_k|/\beta)^p} 
\end{equation}
While $\beta>0$ controls the variance of $c_k$, the value of $0 < p < \infty$ determines the appearance of $c$ in terms its sparsity (with smaller values of $p$ resulting in ``heavier" tails of the corresponding probability density function (pdf), thereby allowing larger values of $c_k$ to be occasionally drawn). In particular, the choice of $p=1$ results in a Laplacian distribution, which is commonly used to describe the behavior of sparse representation coefficients~\cite{Donoho95, Chen01}. Subsequently, the final expression for the MAP estimate becomes
\begin{equation}\label{E8}
c_{MAP} = \argmax_c \prod_{\xx_i=1}^{NM} \left\{ \frac{e^{-({\bf A}[c])_i} ({\bf A}[c])_i^{g(\xx_i)}}{g(\xx_i)!}\right\} \prod_{k \in \mathcal{I}} \left\{ e^{-(|c_k|/\beta)^{p}} \right\}.
\end{equation}

It is conventional to convert the maximization problem (\ref{E8}) into a minimization problem through applying the $\log$-transform to the posterior probability in (\ref{E8}), followed by inverting the sign of the expression thus obtained. In this case, the MAP estimate can be re-expressed as given by
\begin{align}\label{E9}
c_{MAP} &= \argmin_c \left\{ E(c) \right\}, \notag \\ 
E(c) = \langle \mathbf{1}, {\bf A}[c] \rangle - \langle g , &\log({\bf A}[c]) \rangle + \gamma \, \|c\|_p^p, \quad 
\gamma \triangleq 1/\beta^p,
\end{align}
where $\mathbf{1}$ stands for an $N \times M$ matrix of ones, $\langle \, \cdot \,, \, \cdot \, \rangle$ stands for the standard inner product in $\mathbb{R}^{N \times M}$ and $\|c\|_p^p = \displaystyle\sum_{k \in \mathcal{I}} |c_k|^p$ is the $\ell_p$-norm of the representation coefficients. 

It should be noted that the likelihood model of (\ref{E6}) interprets the true image value $f(\xx_i)$ as the mean value of the corresponding random observation $g(\xx_i)$. Moreover, since in the case of the Poisson distribution, the first and the second moments of the distribution are equal, the values $f(\xx_i)$ should be assumed to be nonnegative. The latter assumption restricts the domain of $E(c)$ in (\ref{E9}) to be now defined as
\begin{equation}\label{dom}
{\bf dom} \ E = \left\{ c \in \ell_2(\mathcal{I}) \mid {\bf \Phi}[c] \succeq 0 \right\}.
\end{equation}

Note that $E(c)$ is convex over ${\bf dom} \ E$. Moreover, it is strictly convex over ${\bf dom} \ E$ if $p>1$ {\em or} $g \succ 0$ and {\bf A} is non-degenerate (albeit, possibly, ill-conditioned). In the latter case, the functional $E(c)$ in (\ref{E9}) admits a unique minimizer (note that the set ${\bf dom} \ E$ is convex) \cite[Ch. 11]{Boyd04}, which can be found by any algorithm which is guaranteed to converge to a stationary point of $E(c)$. On the other hand, when either $g$ is not strictly positive or $\Aa$ possesses a non-trivial null-space, the convexity of $E(c)$ is not strict. In such a situation, the uniqueness of a minimizer of $E(c)$ for $p=1$ (which is the case of primary concern in this paper) can be re-established by replacing the $\ell_1$-norm in (\ref{E9}) by its smooth relaxation, e.g., by $\langle {\bf 1}, \sqrt{c^2 + \epsilon} \rangle$, for some $\epsilon \ll 1$ \cite{Figueiredo03}. It should be noted, however, that as long as practical aspect of the image reconstruction is concerned, it has been observed that, for the case of the method proposed in this paper, the above relaxation seems to be unnecessary even in the case of degenerate $\bf H$ and $g \succeq 0$.

It should be noted that the problem of minimizing $E(c)$ over the set defined by (\ref{dom}) does not necessarily have to be formulated as a constrained minimization problem. This is because, for nonnegative data images, the term $- \langle g , \log({\bf A}[c]) \rangle$ in (\ref{E9}) works similar to a log-barrier function which forces the solution to stay within the feasible region defined by the condition ${\bf \Phi}[c] \succeq 0$. In fact, strong theoretical guarantees for the existence and attainability of a unique minimizer of $E(c)$ follow directly from the theory of interior-point methods~\cite[Ch.11]{Boyd04}.  

A possible solution to computing $c_{MAP}$ could be by means of standard optimization techniques, such as gradient-based methods~\cite{Bertsekas99}. It was recently argued, however, that such general purpose tools might be ineffective when applied to the problem at hand, since they do not exploit properly the sparse structure of the desired solution \cite{Elad07}. In the following section, a different approach to finding $c_{MAP}$ is detailed based on the methodology of iterative shrinkage~\cite{Daubechies03, Figueiredo03, Elad07, Blumensath08}.   

\section{Bound optimization and iterative shrinkage}
The non-differentiability of the $\ell_p$-norm for the case $p=1$ (which forms the main focus of this paper) makes it impossible to solve (\ref{E9}) by means of gradient-based optimization methods. Fortunately, an effective solution to the problem can be derived using the majorization-minimization (MM) approach of \cite{Hunter04} (also known as the method of bound optimization~\cite{Figueiredo05}). The first step in applying this method consists in replacing the original problem of minimizing $E(c)$ by a problem of minimizing a {\em surrogate} functional $Q(c, c_0)$ (with $c_0 \in \ell_2(\mathcal{I})$ being an arbitrary set of coefficients). Specifically, let the new functional $Q(c, c_0)$ have the following form
\begin{equation}\label{E10} 
Q(c, c_0) = E(c) + \Psi(c, c_0),
\end{equation}
where
\begin{equation}\label{E11} 
\Psi(c, c_0) = \langle g , \log \left( {\bf A}[c] \slash {\bf A}[c_0] \right) \rangle -  \langle {\bf A}^\ast\left[ g \slash {\bf A}[c_0] \right], c-c_0 \rangle_{\ell_2(\mathcal{I})} + \frac{\mu}{2} \|c - c_0 \|_2^2, 
\end{equation}
with the ``slash" to be interpreted as an element-wise division, ${\bf A}^\ast$ being the adjoint of $\bf A$, and $\langle \cdot, \cdot \rangle_{\ell_2(\mathcal{I})}$ standing for the inner product in $\ell_2(\mathcal{I})$. It goes without saying that the surrogate functional in (\ref{E11}) has to obey a number of critical constraints  which will be detailed in Section~\ref{CA}.

To minimize $Q(c, c_0)$ with respect to $c$, its gradient is computed first according to 
\begin{align}\label{E12}
\nabla Q(c, c_0) = \Aa^\ast[1] - {\bf A}^\ast\left[ g \slash {\bf A}[c] \right] + p \, \gamma \, |c|^{p-1} {\rm sign}(c) + \\ \notag
+ {\bf A}^\ast\left[ g \slash {\bf A}[c] \right] - {\bf A}^\ast\left[ g \slash {\bf A}[c_0] \right] + \mu (c - c_0). 
\end{align}
Subsequently, canceling the similar terms with opposite signs and equating the gradient to zero yields
\begin{equation}\label{E13}
c + \frac{p \, \gamma}{\mu} |c|^{p-1} {\rm sign}(c) = c_0 + \frac{1}{\mu}  {\bf A}^\ast\left[ g \slash {\bf A}[c_0] - 1 \right].
\end{equation}
It should be noted that the above analysis is relevant to the differentiable case of $p>1$. For the case of $p=1$, the first order optimality condition (\ref{E13}) has to be replaced by its more general version requiring
\begin{equation}\label{E13a}
c + \frac{\gamma}{\mu} \partial \|c\|_1 \ni c_0 + \frac{1}{\mu}  {\bf A}^\ast\left[ g \slash {\bf A}[c_0] - 1 \right],
\end{equation}
where $\partial \|c\|_1$ denotes the subdifferential of $\|c\|_1$. Remarkably, both (\ref{E13}) and (\ref{E13a}) have the same closed form solution. Particularly, given the {\em inverse} $\mathcal{S}_{p, \gamma, \mu}(c)$ of the function on the left-hand side of (\ref{E13}), the optimal $c$ that minimizes $Q(c,c_0)$ is given by
\begin{equation}\label{E14}
c = \mathcal{S}_{p, \gamma, \mu} \left( c_0 + \frac{1}{\mu}  {\bf A}^\ast\left[  \frac{g - \Aa[c_0]}{\Aa[c_0]}  \right] \right).
\end{equation}

The functional in (\ref{E14}) is analogous to the functional used in the existing iterative shrinkage methods, which have been derived under the assumption of Gaussian noises~\cite{Daubechies03, Figueiredo03, Elad07, Blumensath08}. Just like in the latter studies, we propose to find a minimizer of the original functional in (\ref{E9}) iteratively, {\it viz.} 
\begin{equation}\label{E15}
c_{t+1} = \mathcal{S}_{p, \gamma, \mu} \left( c_t + \frac{1}{\mu}  {\bf A}^\ast\left[ \frac{g - \Aa[c_t]}{\Aa[c_t]}  \right] \right),
\end{equation} 
where $t$ stands for the iteration index.
The convergence of the above iteration scheme to a global minimizer of (\ref{E9}) is proven in the next section. Before turning to the proof, we note that setting $p=1$ seems to be a reasonable choice in practice, since it guarantees the convexity of $E(c)$ on one hand, and leads to a sparse solution on the other. It is worthwhile noting that, in this case, the function $\mathcal{S}_{p, \gamma, \mu}$ becomes the notorious {\em soft thresholding} rule \cite{Donoho95}, which is defined as
\begin{equation}\label{E16}
\mathcal{S}_{1, \gamma, \mu}(c) = 
\begin{cases}
(|c| - \gamma/\mu) \, {\rm sign(c)}, \quad &{\rm if} \,\,  |c| \geq \gamma/\mu \\ 
0, \quad &{\rm otherwise}.
\end{cases}
\end{equation}

Finally, it is worthwhile noting that the computational cost in (\ref{E15}) is mainly determined by the cost of applying the composition transform ${\bf A} = {\bf H} \cdot {\bf \Phi}$ and its adjoint. While the convolution operator can be computed efficiently, e.g., by means of the fast Fourier transform \cite{Gonzales02}, the application of $\bf \Phi$ and ${\bf \Phi}^\ast$ depends on the type of transformation in use. Fortunately, most of the relevant transforms (such as wavelet \cite{Mallat99}, ridgelet \cite{Candes99}, and curvelet \cite{Candes06} transforms) admit computationally efficient implementations, which are typically of a logarithmic complexity at most. A formal comparison between the computational efficiencies of the proposed and reference methods is given in Section V.

\section{Convergence analysis}\label{CA}
To demonstrate that the proposed algorithm constitutes a viable alternative to traditional approaches, its convergence properties need to be analyzed next. As was shown in~\cite{Hunter04}, the convergence can be guaranteed provided that the surrogate functional $Q(c,c_t)$ satisfies the following conditions:
\begin{align}\label{E17}
1)~&Q(c,c_t) \geq E(c), \,\, \forall \, c \in {\bf dom} \, E \notag \\
2)~&Q(c_t,c_t) = E(c_t)
\end{align}
with $c_t$ being an arbitrary but fixed reference point in ${\bf dom} \, E$.

Before verifying the above conditions, it is instructive to show first that minimizing $Q(c,c_t)$ with respect to $c$ yields a reduction in the value of the original functional $E(c)$. To this end, let $c_{t+1}$ denote a minimizer of $Q(c,c_t)$, i.e. $c_{t+1} = \argmin_{c} Q(c,c_t)$. For such a $c_{t+1}$ and $Q(c,c_t)$ obeying (\ref{E17}), one then obtains 
\begin{equation}\label{E18}
E(c_{t+1}) = Q(c_{t+1},c_t) + E(c_{t+1}) - Q(c_{t+1},c_t) \leq Q(c_t,c_t) + E(c_t) - Q(c_t,c_t) = E(c_t),
\end{equation}
resulting in
\begin{equation}\label{E19a}
E(c_{t+1}) \leq E(c_t).
\end{equation}
Thus, the bound-optimization algorithm (\ref{E15}) guarantees a reduction in the value of $E(c_t)$ at each iteration provided $Q(c,c_t)$ obeys the conditions in (\ref{E17}). It should be noted, however, that for the value of $E(c_{t+1})$ to decrease, it is sufficient to require that $Q(c,c_t) \geq E(c)$ {\em only} at $c=c_{t+1}$. The latter seems to be much less restrictive condition as compared to Condition 1 in (\ref{E17}). (We will take advantage of this fact later in this section.)

The fact that Condition 2 in (\ref{E17}) is satisfied by the proposed surrogate functional $Q(c,c_t)$ can be verified by direct substitutions. Moreover, it is important to point out that the function $\Psi$ in (\ref{E10}) obeys
\begin{equation}\label{E19b}
\Psi(c_t,c_t) = 0 \quad {and} \quad \nabla \Psi(c,c_t) \big|_{c=c_t} = 0,
\end{equation} 
which suggests that $\Psi(c,c_t)$ has an extremum at $c=c_t$ that is equal to zero. Consequently, if $\Psi(c,c_t)$ is a convex function, Condition 1 will be automatically fulfilled. The convexity, on the other hand, can be determined from the properties of the Hessian operator corresponding to $\Psi(c,c_t)$, which can be shown to be equal to
\begin{equation}\label{E20}
\nabla^2 \Psi(c) = \mu \, {\bf I} - \Aa^\ast \, {\rm diag}\left( \frac{g}{(\Aa [c])^2} \right) \, \Aa.
\end{equation}
Note that, in the expression above, $\bf I$ stands for the identity operator, while the diagonal operator $ {\rm diag}\left( g \slash (\Aa [c])^2 \right)$ is defined in the standard way as 
\begin{equation}
{\rm diag}\left( \frac{g}{(\Aa [c])^2} \right)[y] \equiv \frac{g}{(\Aa [c])^2} \cdot y,
\end{equation}
for any arbitrary $y \in \mathbb{R}^{N\times M}$. Provided the elements of $g \slash (\Aa [c])^2$ are bounded and $\Aa$ is compact, the Hessian in (\ref{E20}) is necessarily a self-adjoint and compact operator. Moreover, the minimial eigenvalue $\lambda_{min}$ of the Hessian can be shown to be bounded by
\begin{equation}\label{E21}
\lambda_{min} \geq \mu - \left\| g \slash (\Aa[c])^2  \right\|_\infty \lambda_{max}(\Aa^\ast \Aa),
\end{equation} 
where $\|\cdot\|_\infty$ stands for the supremum norm and $\lambda_{max}(\Aa^\ast \Aa)$ denotes the maximum eigenvalue of $\Aa^\ast \Aa$. Thus, as long as 
\begin{equation}\label{E22}
\mu \geq \left\| g \slash (\Aa[c])^2  \right\|_\infty \lambda_{max}(\Aa^\ast \Aa),
\end{equation}
the Hessian $\nabla^2 \Psi(c)$ is positive definite, in which case $\Psi(c,c_t)$ is convex, and hence Condition 1 of (\ref{E17}) is satisfied. It should be pointed that the finiteness of $\mu$ is always guaranteed by the fact that $c \in {\bf dom} \, E$. The value of $\mu$ in (\ref{E22}), however, is defined as a function of $c$, and therefore it would be quite problematic (if possible at all) to determine $\mu$ from (\ref{E22}), had we decided to do so. In this sense, the condition (\ref{E22}) should be regarded as merely a ``proof of existence".

It turns out that in practical cases there are much simpler means to determine a value of $\mu$ that guarantees a reduction in the value of $E(c)$ with respect to $E(c_t)$. First, we note that according to (\ref{E18}), it is sufficient to find a $\mu_t$ that results in\footnote{The subscript $t$ is added intentionally in $\mu_t$ to indicate that its value can be iteration-dependent.} 
\begin{equation}\label{E23}
\frac{\mu_t}{2} \|c_{t+1} - c_t \|_2^2 \geq \langle {\bf A}^\ast\left[ g \slash {\bf A}[c_t] \right], c_{t+1}-c_t \rangle_{\ell_2(\mathcal{I})} - \langle g , \log \left( {\bf A}[c_{t+1}] \slash {\bf A}[c_t] \right) \rangle,
\end{equation}
thereby guaranteeing that $\Psi(c_{t+1},c_t) \geq 0$, and therefore $E(c_{t+1}) \leq E(c_t)$. This implies the following practical way to find an acceptable $\mu_t$. Let the right-hand side of (\ref{E23}) be denoted by $F(c_{t+1},c_t)$, which is a computable quantity provided the values of $c_t$ and $c_{t+1}$. Consequently, a suitable value of $\mu_t$ can be found using the following algorithm. 

\begin{algorithm}
\caption{Finding a suitable scaling parameter $\mu_t$} 
\begin{algorithmic}[1]
\STATE {\textbf{Preset:}} {$\nu = 1$, $0 < \alpha < 1$}
\STATE {\textbf{Compute:}} {$c_{t+1} = \mathcal{S}_{1,\gamma,1} \left(c_t + \Aa^\ast \left[  g/\Aa[c_t] - 1 \right] \right)$}
\IF {$ \|c_t - c_{t+1}\|_2^2 \geq 2 \, F(c_{t+1},c_t)$}
\WHILE {$ \nu \, \|c_t - c_{t+1}\|_2^2 \geq 2 \, F(c_{t+1},c_t)$}
\STATE {$\nu \Leftarrow \alpha \, \nu$}
\STATE {$c_{t+1} = \mathcal{S}_{1,\gamma,\nu} \left(c_t + \frac{1}{\nu} \Aa^\ast \left[  g/\Aa[c_t] - 1 \right] \right)$}
\ENDWHILE
\STATE {$\mu_t = \nu \slash \alpha$}
\ELSE
\WHILE {$ \nu \, \|c_t - c_{t+1}\|_2^2 < 2 \, F(c_{t+1},c_t)$}
\STATE {$\nu \Leftarrow \nu \slash \alpha$}
\STATE {$c_{t+1} = \mathcal{S}_{1,\gamma,\nu} \left(c_t + \frac{1}{\nu} \Aa^\ast \left[  g/\Aa[c_t] - 1 \right] \right)$}
\ENDWHILE
\STATE {$\mu_t = \nu$}
\ENDIF
\RETURN {$\mu_t$}
\end{algorithmic}
\end{algorithm} 

Algorithm 1 has been designed to find a {\em minimal possible} $\mu$ (with the accuracy of $\log{\alpha}$ in the logarithmic scale) that guarantees that\footnote{The accuracy can be improved via choosing $\beta$ to be close to 1, which has a drawback of slowing down the convergence.} $E(c_{t+1}) \leq E(c_t)$. It should be emphasized that the proof of existence in (\ref{E22}) assures that an acceptable value of $\mu_t$ can always be found. The form of the shrinkage operator in (\ref{E15}) suggests that smaller values of $\mu_t$ result in more substantial shrinkage, which in turn leads to more sizable changes in $c_{t+1}$ with respect to $c_t$. The idea of maximizing the effect of shrinkage through minimizing the value of $\mu_t$ lies in the heart of the method proposed in~\cite{Elad06}. In the present case, the minimality of $\mu_t$ is guaranteed by the design of Algorithm 1, whose only downside is in the extra calculations required. In practical scenarios, however, when one is working with a pool of images of similar nature, Algorithm 1 can be excluded by predefining the value of $\mu$. This can be done, for example, by applying Algorithm 1 to several images and determining a fixed value of $\mu$ that satisfies (\ref{E23}).

\section{Results}
\subsection{Reference methods}
In the experimental part of this paper, the proposed method is referred to as {\em Poisson Iterative Shrinkage} (PIS). As was argued earlier, PIS is conceptually parallel to the iterative shrinkage method developed in~\cite{Daubechies03, Figueiredo03} under the assumption of Gaussian noises, in which case the iterative shrinkage (referred to below as {\em Gaussian Iterative Shrinkage} (GIS)) has the form of
\begin{equation}\label{GIS}
c_{t+1} = \mathcal{S}_{p, \gamma, \mu} \left( c_t + \frac{1}{\mu}  {\bf A}^\ast\left[ g - \Aa[c_t] \right] \right)
\end{equation}
with $\gamma = \sigma^2/\beta$ where $\beta$ is the bandwidth parameter of the GG pdf in (\ref{E7}) and $\sigma^2$ is the variance of the i.i.d. Gaussian noise contaminating the (blurred) measurements of the true image. The GIS algorithm is known to converge to the global minimizer of a MAP criterion provided $\mu > \|\Aa^\ast \Aa\|$ \cite{Daubechies03}. 

Another reference method used in the comparison is the {\em Richardson-Lucy} (RL) algorithm, which is represented by the following iterative scheme 
\begin{equation}\label{RL}
f_{t+1} =  f_t \ {\bf H}^\ast \left[ \frac{g}{ {\bf H}[f_t] }  \right]. 
\end{equation}
Note that, as opposed to the GIS and PIS methods, the iterations in RL are performed on an estimated image $f_t$ rather on the coefficients of its representation in a basis/frame. The iteration procedure in (\ref{RL}) is derived from a maximum-likelihood (ML) model under the assumption of Poisson noises. ML estimators, however, are known to result in degraded performance in the case of poorly conditioned $\bf H$. To alleviate this deficiency, \cite{Dey06} proposed to use the MAP framework to regularize the convergence in (\ref{RL}). This method assumes the true $f$ to be in the space of bounded-variation images and, thus, the resulting iterative scheme minimizes the total variation (TV) norm of the estimate of $f$. Using the methodology of RL, the above minimization can be performed via
\begin{equation}\label{RLTV}
f_{t+1} =   \frac{f_t}{1 - \gamma \, {\rm div} \left( \frac{\nabla f_t}{\|\nabla f_t\|} \right) } \ {\bf H}^\ast \left[ \frac{g}{ {\bf H}[f_t] }  \right],
\end{equation}
where $\gamma$ is a regularization parameter, which is to obey $1 - \gamma \, {\rm div} \left( \frac{\nabla f_t}{\|\nabla f_t\|} \right) \succ 0$ to guarantee the reconstruction to be non-negative. Below, the algorithm of (\ref{RLTV}) is referred to as the {\rm Richardson-Lucy total variation} (RLTV) method. 

A different reference method that also takes advantage the TV regularization is \cite{Setzer09}. This method is based on minimizing the same objective function as RTLV, subject to a non-negativity constraint on $f$. The solution in \cite{Setzer09} uses the variable splitting technique of \cite{Goldstein09}, which allows reducing the minimization problem to a few simpler subproblems. In what follows, the method is referred to as PIDSplit+ (which stands for Poisson image deconvolution by variable splitting with a positivity constraint).

Another reference approach used in the present study is that of \cite{Dupe08}. Similarly to the method proposed in the present paper, \cite{Dupe08} assumes $f$ to be sparsely representable in the domain of a linear transform. Subsequently, the algorithm is initialized by applying a VST (namely, the Anscombe transform) to the data image $g$, followed by recovering $f$ as a solution to a standard $\ell_2 - \ell_1$ minimization problem. In the present study, the above method has been implemented using a publicly available code (see {\tt http://www.greyc.ensicaen.fr/$\sim$fdupe/}). Due to the fact that \cite{Dupe08} incorporates a VST with the theory sparse representations, in the discussion that follows, this method is referred to as VSTSR.

The last reference method used in our comparative study is the one described in \cite{Harmany09}. The method employes the sparsity assumption in the domain of an orthogonal wavelet transform. Subsequently, \cite{Harmany09} solves the minimization problem of (\ref{E9}) by minimizing a sequence of quadratic approximations to a $\log$ penalty function. Following \cite{Harmany09}, the above method is referred to as sparse Poisson intensity reconstruction algorithm (SPIRAL).   

Since RL, RLTV and PIDSplit+ tend to become unstable in the case of poorly conditioned $\Hh$, a common practice is to terminate their execution after a predefined number of iterations. In our experiments, their termination was performed at the point where the {\em normalized mean-squared error} (NMSE) (defined below) reached its minimum value. Needless to say that such termination is only possible under the conditions of controlled simulation studies, when the original images are known. In practical scenarios, however, the NMSE-optimal termination is generally impossible, which suggests that real-life reconstructions obtained with RL, RLTV and PIDSplit+ may be actually worse than the reconstructions demonstrated in the present paper. The proposed PIS algorithm, on the other hand, remains stable in the course of its convergence, which makes it possible to terminate the algorithm simply after a relative change in the value of $E(c)$ drops below a predefined threshold (e.g., $10^{-6}$). 

It is worthwhile noting that the reference methods above have been derived using different statistical approaches and assumptions. The motivation behind choosing these methods has been to compare the proposed method to the approaches based on the Poisson model (i.e. RL, RLTV, PIDSplit+, VSTSR, SPIRAL) as well as to those exploiting the idea of iterative shrinkage (i.e. GIS). The main properties of all the reconstruction methods under consideration are summarized in Table~\ref{T1}.

\begin{table}[here]
\caption{Properties of the reconstruction methods under comparison}
\label{T1} 
\begin{center}
\begin{tabular}{ | c | c | c | c | c | }
\hline
& {\bf Statistical framework} & {\bf Prior model}  & {\bf Noise model} \\ \hline\hline
{\bf RL}		    	&   		ML    	& 		N/A 						& 		Poisson \\ \hline
{\bf RLTV}	    	&  		MAP  	& 		BV image $f$ 			& 		Poisson \\ \hline
{\bf PIDSplit+}	&   		MAP  	& 		BV image $f$ 			& 		Poisson \\ \hline
{\bf VSTSR}	    	&   		MAP  	& 		GG coefficients $c$ 		& 		Poisson \\ \hline
{\bf SPIRAL}	  	&   		MAP  	& 		Laplacian coefficients $c$ 	& 		Poisson \\ \hline
{\bf GIS}		    	&	   	MAP	&		GG coefficients $c$ 		&		Gaussian  \\ \hline
{\bf PIS}		    	&	    	MAP	&		GG coefficients $c$ 		&		Poisson  \\ \hline
\end{tabular}
\end{center}
\end{table}

\subsection{Computational complexity}
The most computationally demanding operations used in the proposed and reference methods are associated with the convolution $\Hh$ and frame ${\bf \Phi}$ operators. The convolution operator $\Hh$ and its adjoint can be efficiently implemented by using, e.g., the FFT algorithm, which requires $n \log n$ MAC operations, with $n = N M$ representing the total number of pixels. Since the implementation of convolution and its related complexity depend on $\Hh$ itself, let $C(n)$ represent the total number of MAC operations required by computing $\Hh$ and $\Hh^\ast$. In a similar manner, let $R(n)$ represent the total number of MAC operations required by applying the frame operator $\bf \Phi$ and its adjoint. (Thus, for example, $R(n) \sim \mathcal{O}(n)$ in the case of ${\bf \Phi}^\ast$ being an orthogonal wavelet transform). Consistent with the notations above,  Table~\ref{T1a} summarizes the computational complexities required by {\em one iteration} of the proposed and reference methods. Needless to say, the overall complexity of the above methods will depend on the number of iterations required till their convergence. These numbers will be provided below for specific examples of image reconstruction.

\begin{table}[top]
\caption{Computational complexity of the reconstruction methods under comparison}
\label{T1a} 
\begin{center}
\begin{tabular}{ | c | c | c |}
\hline
& {\bf Computational operations per iteration}  \\ \hline\hline
{\bf PIS}		    	&	       $2C(n) + 2R(n) + \mathcal{O}(n)$	\\ \hline
{\bf GIS}		    	&	       $2C(n) + 2R(n) + \mathcal{O}(n)$  \\ \hline
{\bf RL}		    	&   		$2C(n) + \mathcal{O}(n)$       					\\ \hline
{\bf RLTV}	    	&   		$2C(n) + \mathcal{O}(n)$       					\\ \hline
{\bf PIDSplit+}	&   		$4C(n) + \mathcal{O}(n)$ 	    					\\ \hline
{\bf VSTSR}	  	&   		$2C(n) + \mathcal{O}(R(n))$ 	    				\\ \hline
{\bf SPIRAL}	  	&   		$2C(n) + \mathcal{O}(n)$ 	    					\\ \hline
\end{tabular}
\end{center}
\end{table}

\subsection{Comparison measures}
The algorithms listed in Table~\ref{T1} have been compared in terms of the NMSE defined as follows. Let $f$ be a true image and $\tilde{f}$ be an estimate of $f$. Then, the NMSE can be defined as
\begin{equation}\label{NMSE}
{\rm NMSE} = \mathcal{E} \left\{ \frac{\| f - \tilde{f} \|_F^2}{\| f \|_F^2} \right\},
\end{equation}
with $\| \cdot \|_F$ being the Frobenius matrix norm, and $\mathcal{E}$ being the operator of expectation. In the current study, the latter is approximated by sample mean based on the results of 200 independent trials.

It has been recently argued that the NMSE may not be an optimal comparison measure as long as human visual perception is concerned. For this reason, the NMSE-based comparison has been complimented by comparing the reconstruction algorithms in terms of the SSIM index as suggested in~\cite{Wang04}.

\subsection{Sparse deconvolution}
Our first example is concerned with the problem of sparse deconvolution\footnote{In this case, the image $f$ is identified with its coefficients $c$, i.e. $f \equiv c$.}, in which case $\Aa \equiv {\bf H}$. In this task, our main intension is to demonstrate the importance of regularization and correct modeling of noise for successful reconstruction of $f$ in (\ref{E1}). Consequently, in this subsection, the performance of the PIS algorithm is compared to those of RL and GIS.

\begin{figure}[htbp]
\centering
\includegraphics[width=5in]{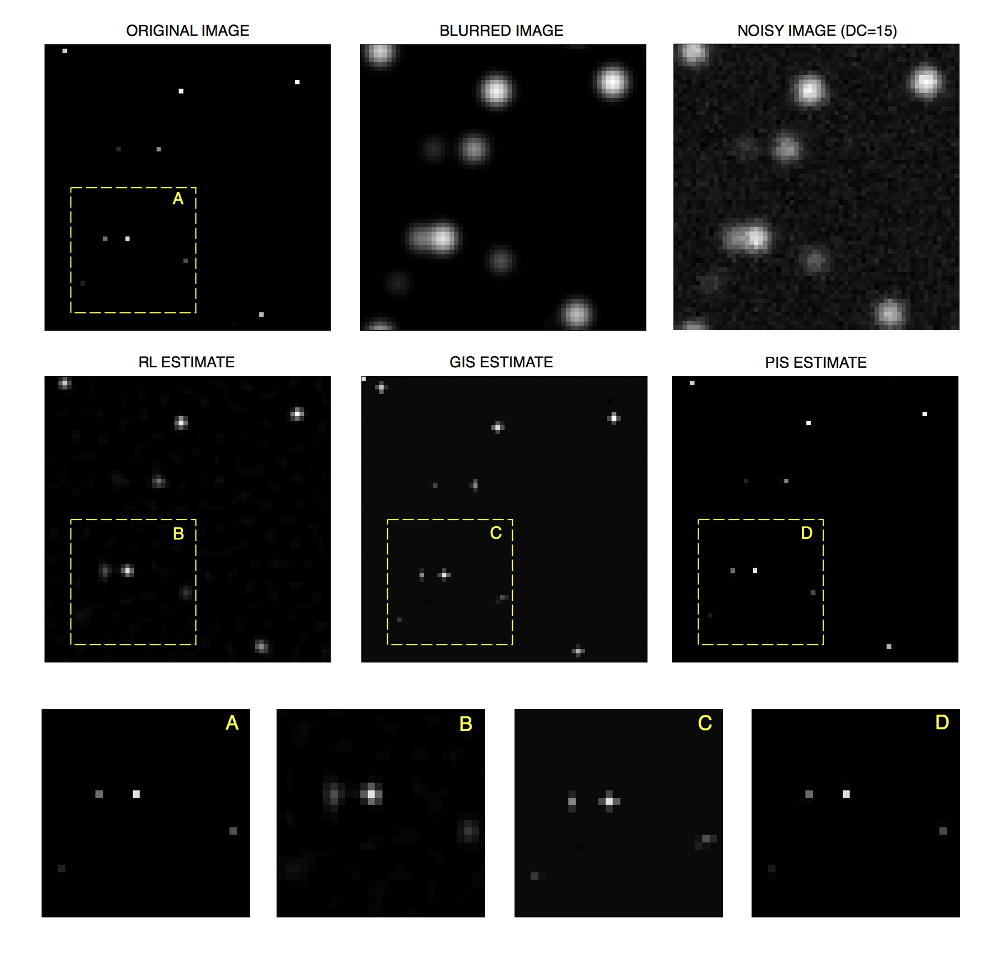} 
\caption{(First row of subplots) Original image, blurred image, and noisy image (SNR=17); (Second row of subplots) RL reconstruction, GIS reconstruction, and PIS reconstruction; (Third row of subplots) Zoomed segments of the original and reconstructed images as indicated by the dashed rectangles.}
\label{F1}
\end{figure}

The assumption of sparsity suggests that $f$ consists of a small number of bright sources scattered over a black background. An example of such an image is shown in the upper-right subplot of Fig.~\ref{F1}, where the non-zero samples of $f$ have been generated by taking the absolute value of an i.i.d. Gaussian random variables. 

\begin{figure}[htbp]
\centering
\includegraphics[width=5in]{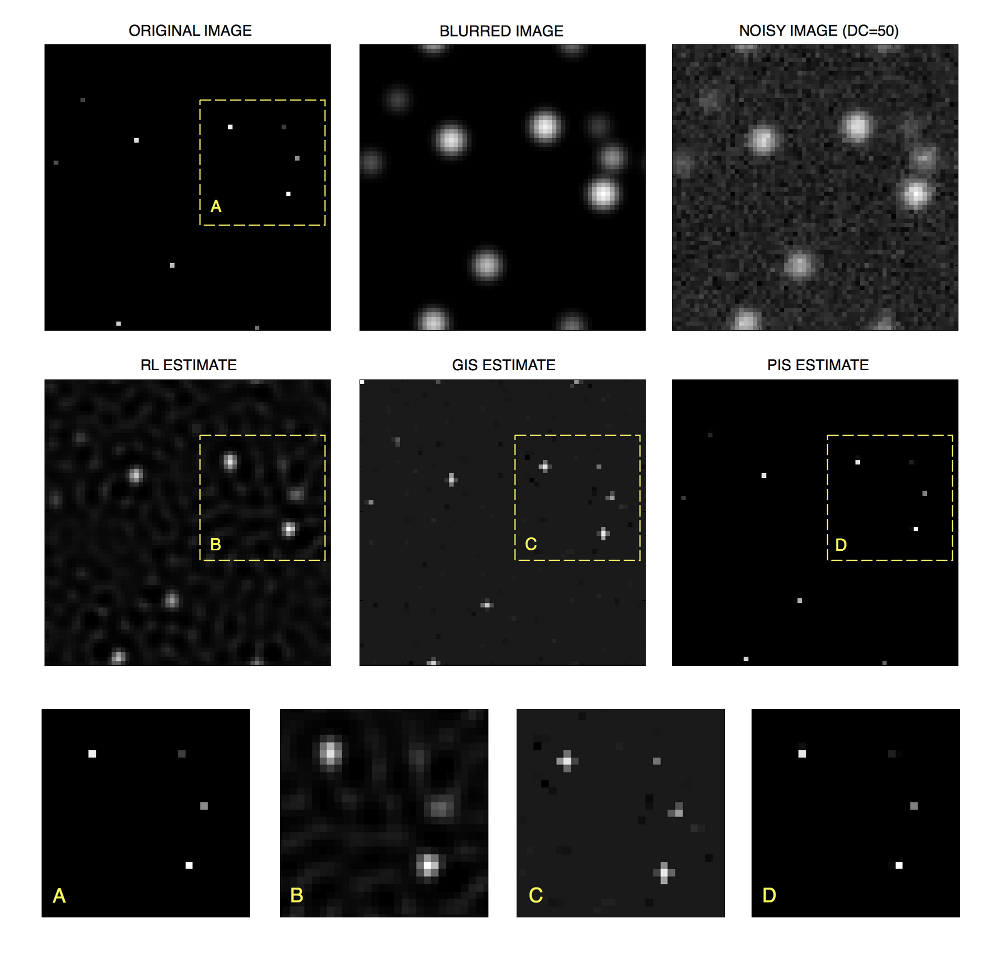} 
\caption{(First row of subplots) Original image, blurred image, and noisy image (SNR=5.1); (Second row of subplots) RL reconstruction, GIS reconstruction, and PIS reconstruction; (Third row of subplots) Zoomed segments of the original and reconstructed images as indicated by the dashed rectangles.}
\label{F2}
\end{figure}

The blurring artifact was simulated by convolving the test images with an isotropic Gaussian kernel $h(\xx)$ whose -3 dB cut-off frequency was set to be equal to $0.2 \pi$. As the next step, the resulting images were contaminated by Poisson noise. In this regard, it should be noted that the level of Poisson noise is defined by the corresponding value of the blurred image $\Hh [f]$. Since real-life images are always contaminated by background noises, the minimum value of $\Hh [f]$ should be strictly positive. In this study, a set of three different minimum (background) values, namely 15, 30, and 50, were used. Note that since the variance of Poisson noise is equal to its mean value, higher background values will result in more severe noises. In this case, it seems to be reasonable to define the SNR as a ratio of the maximum value of $\Hh [f]$ (i.e. 255) to its background value. According to this definition, the SNR values used in the present study were 17, 8.5, and 5.1. Two examples of simulated data images for SNR=17 and SNR=5.1 are shown in the upper-left subplots of Fig.~\ref{F1} and Fig.~\ref{F2}, respectively.

In the case of low-pass blurs, the only possibility for $\Hh[f]$ to generate a constant background is to require $f(\xx) \neq 0$ for all $\xx \in \Omega$. This obviously contradicts the assumption on $f$ to be sparse. To alleviate this deficiency, we suggest to modify the image formation model via replacing $\Hh$ and $f$ by $\tilde{\Hh} \triangleq [\Hh \,\,\,\, 1]$ and $\tilde{f} \triangleq [f \,\,\,\, f_0]^T $, respectively, where $f \in \mathbf{R}^{K\times L}$ is assumed to be sparse, and $f_0 \in \mathbb{R}_+\backslash \{0\}$ is a positive scalar defining the background level. In this case, the image formation model of (\ref{E1}) can be redefined as  
\begin{equation}\label{DC}
g = \mathcal{P}\left\{ \tilde{\Hh}[\tilde{f}] \right\} = \mathcal{P}\left\{ \Hh[f] + f_0 \right\}.
\end{equation} 
Consequently, the reconstruction is applied with $\tilde{\Hh}$ to recover $\tilde{f}$, in which case the true image is considered to be equal to the sum $f+f_0$.

\begin{figure}[htbp]
\centering
\includegraphics[width=5.5in]{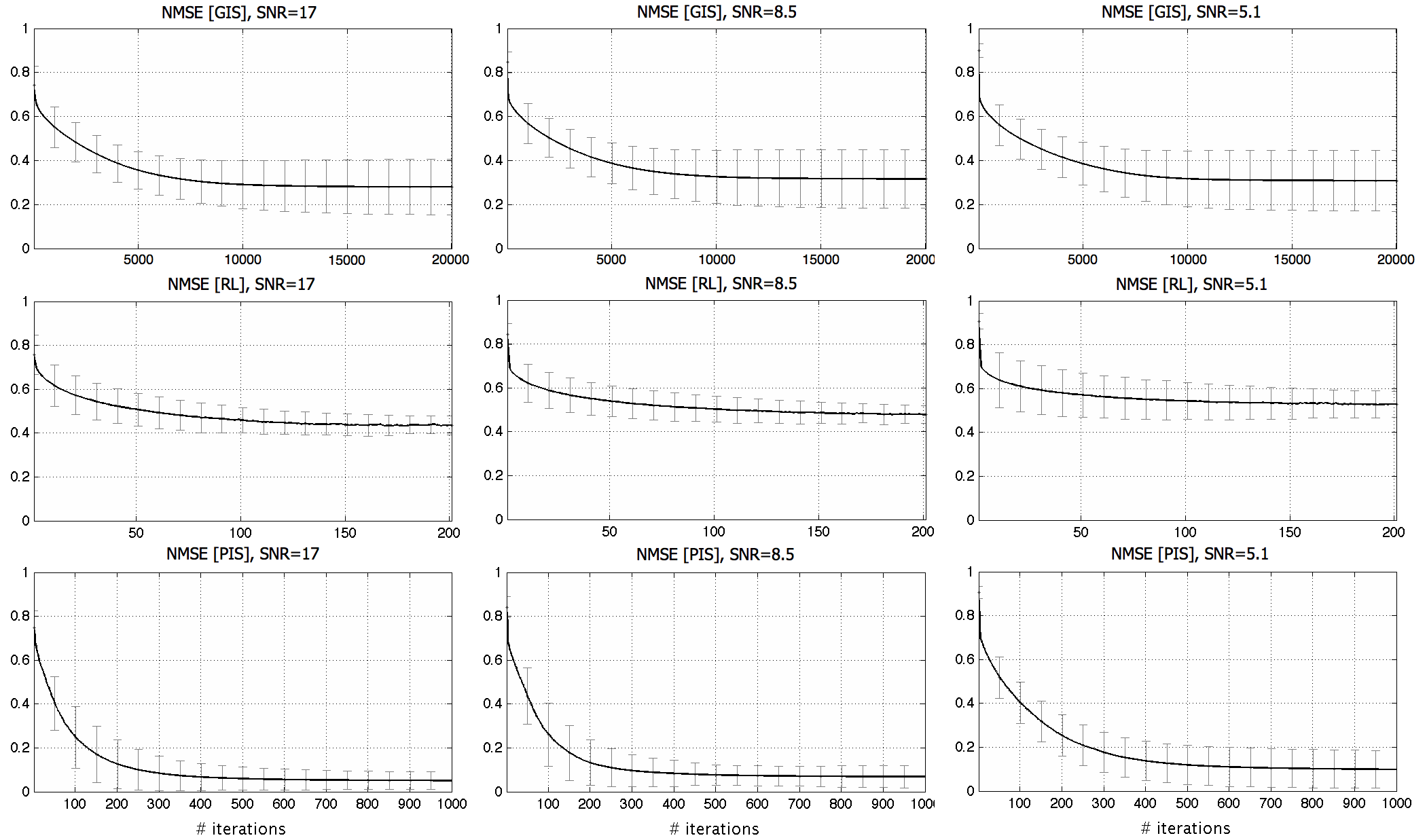}
\caption{(Upper row of subplots) The NMSE of GIS reconstruction as a function of the number of iterations for SNR=17, 8.5 and 5.1; (Middle row of subplots) The NMSE of RL reconstruction as a function of the number of iterations for SNR=17, 8.5 and 5.1; (Lower row of subplots) The NMSE of PIS reconstruction as a function of the number of iterations for SNR=17, 8.5 and 5.1.}
\label{F3}
\end{figure}

The above model adjustment was applied only for the cases of GIS and PIS reconstruction, which are based on the sparsity assumption. The regularization parameter $\gamma$ in (\ref{E9}) and (\ref{GIS}) was defined to be $1/\beta$ and $\sigma^2/\beta$, respectively, with $\beta=4.5$ and $\sigma^2$ equal to the sample variance of the background noise. In the case of GIS, the parameter $\mu$ was set to be equal to $1.1\cdot \|\tilde{\Hh}^\ast \tilde{\Hh}\|$, while in the case of PIS it was chosen using Algorithm 1 with $\alpha=0.8$. 

Typical reconstruction results are demonstrated in Fig.~\ref{F1} and Fig.~\ref{F2} for SNR=17 and SNR=5.1, correspondingly. In particular, the middle row of subplots of the figures show the reconstructions obtained by the (from left to right) RL, GIS, and PIS algorithms. For the convenience of the reader, the  bottom row of subplots in Fig.~\ref{F1} and Fig.~\ref{F2} show zoomed fragments of the original and recovered images as indicated by the dashed boxes and letters A, B, C, and D. One can see that PIS outperforms all the reference methods in terms of the resolution improvement and noise reduction.

A quantitative comparison of the reconstruction algorithms is presented in Fig.~\ref{F3}, which shows the NMSE as a function of the number of iterations for (from up to down) GIS, RL, and PIS, and for different values of SNR, namely (from left to right) 17, 8.5, and 5.1. It should be noted that each value of the NMSE in Fig.~\ref{F3} is a result of averaging the errors obtained in a series of independent trials, where both the true images and noises were drawn randomly. 

Observing Fig.~\ref{F3}, one can see that PIS results in considerably lower values of the NMSE as compared to the reference methods. As well, it converges to a steady-state solution after a much smaller number of iterations as compared to the GIS algorithm (i.e. 500 vs. $10^4$). Unlike the RL method, which is non-monotonely convergent in NMSE (which can not be seen in Fig.~\ref{F3} since the method was terminated before the algorithm diverged), the convergence of PIS is monotone in both $E(c)$ and NMSE.

\subsection{Sparse reconstruction}
In the second part of the experimental study, the PIS method was tested in application to the problem of sparse image reconstruction, where the combined operation of image synthesis and blur is represented by $\Aa = \Hh {\bf \Phi}$. In this case, the blur model was defined by the convolution kernel $h[i,j] = (i^2+j^2+1)^{-1}$, with $i,j = -D,\ldots,D$ and $D\in\{2,7\}$~\cite{Elad07}. The frame operator $\bf \Phi$ was defined to describe the translation invariant (TI) wavelet transform corresponding to the Haar wavelet. The number of wavelet resolutions was set to be equal to 4. As in the case with sparse deconvolution, the wavelet frame was extended by adding a constant vector so as to allow the image background to be modeled by a single element of the frame. 

\begin{figure}[top]
\centering
\includegraphics[width=4.2in]{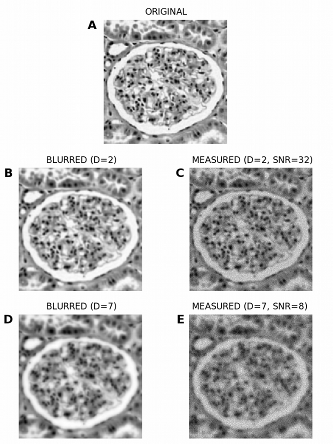}
\caption{(Subplot A) Original image of glomerulus; (Subplot B) Blurred image of glomerulus with D=2; (Subplot C) Blurred and noisy image of glomerulus with D=2 and SNR=32; (Subplot D) Blurred image of glomerulus with D=7; (Subplot E) Blurred and noisy image of glomerulus with D=7 and SNR=8.}
\label{F4}
\end{figure}

\begin{figure}[top]
\centering
\includegraphics[width=4.2in]{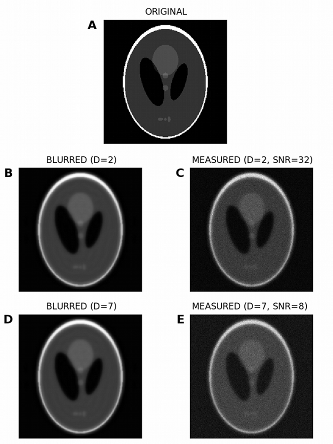}
\caption{(Subplot A) Original image of the Shepp-Logan phantom; (Subplot B) Blurred image of the Shepp-Logan phantom with D=2; (Subplot C) Blurred and noisy image of the Shepp-Logan phantom with D=2 and SNR=32; (Subplot D) Blurred image of the Shepp-Logan phantom with D=7; (Subplot E) Blurred and noisy image of the Shepp-Logan phantom with D=7 and SNR=8.}
\label{F5}
\end{figure}

In this subsection, image reconstructions produced by the proposed and reference methods are tested using a microscopic image of glomerulus and the standard Shepp-Logan phantom, which are shown in Subplots A of Fig.~\ref{F4} and Fig.~\ref{F5}, respectively. Similar to the case of sparse reconstruction, the images have been offset by a constant (background) value to give rise to different values of SNR. In particular, the value was adjusted to result in SNR equal to 32 (moderate noises) and 8 (strong noises). The original, blurred, and contaminated images of the glomerulus and Shepp-Logan phantom are summarized in Fig.~\ref{F4} and Fig.~\ref{F5} for all the tested values of $D$ and SNR. 

\begin{table}[here]
\caption{NMSE and SSIM values of the reconstruction methods under comparison using the Shepp-Logan phantom.}
\label{T2}
\begin{center}
\begin{tabular}{ | l | c | c | c | c | c | c |}
\hline
	 	& \multicolumn{3}{c|}{L=2, SNR=32} & \multicolumn{3}{c|}{L=2, SNR=8} \\ \cline{2-7}
	 				      & NMSE & SSIM & NIT & MNSE & SSIM & NIT\\ \hline
{\bf RL} 	      &  0.101 & 0.68  & 10 & 0.368 & 0.54 & 10  \\ \hline
{\bf RLTV}	    & 0.141 & 0.72 & 10 & 0.400 & 0.71 & 10 \\ \hline
{\bf PIDsplit+}	& 0.123 & 0.85 & 5 & 0.872 & 0.68 & 5 \\ \hline
{\bf VSTSR}	    & 0.189 & 0.65 & 300 & 0.377 & 0.65 & 300 \\ \hline
{\bf SPIRAL}	  & 0.132 & 0.76 & 20,000 & 0.378 & 0.72 & 20,000\\ \hline
{\bf PIS}				& {\bf 0.09} & {\bf 0.88} & 500  & {\bf 0.377} &{\bf 0.84} & 500  \\ \hline
\end{tabular}
\end{center}
\end{table}

\begin{table}[here]
\caption{NMSE and SSIM values of the reconstruction methods under comparison using a glomerulus image; NMSE values appear after multiplication by $10^{2}$.}
\label{T3}
\begin{center}
\begin{tabular}{ | l | c | c | c | c | c | c |}
\hline
	 	& \multicolumn{3}{c|}{L=4, SNR=32} & \multicolumn{3}{c|}{L=7, SNR=8} \\ \cline{2-7}
	 				      & NMSE & SSIM & NIT & MNSE & SSIM & NIT\\ \hline
{\bf RL} 	      & 0.280 & 0.86 & 10 & 0.772 & 0.77 & 10 \\ \hline
{\bf RLTV}	    & 0.275 & 0.87 & 10 & 0.755 & 0.78 & 10 \\ \hline
{\bf PIDsplit+}	& 42.54 & 0.90 & 4 & 47.51 & 0.78 & 4 \\ \hline
{\bf VSTSR}	    & 9.150 & 0.78 & 300 & 9.593 & 0.62 & 300 \\ \hline
{\bf SPIRAL}	  & 0.303 & 0.86 & 50,000 & 0.737 & 0.81 & 50,000 \\ \hline
{\bf PIS}				& {\bf 0.210} & {\bf 0.92} & 400 & {\bf 0.729} & {\bf 0.86} & 400 \\ \hline
\end{tabular}
\end{center}
\end{table}

\begin{figure}[here]
\centering
\includegraphics[width=5in]{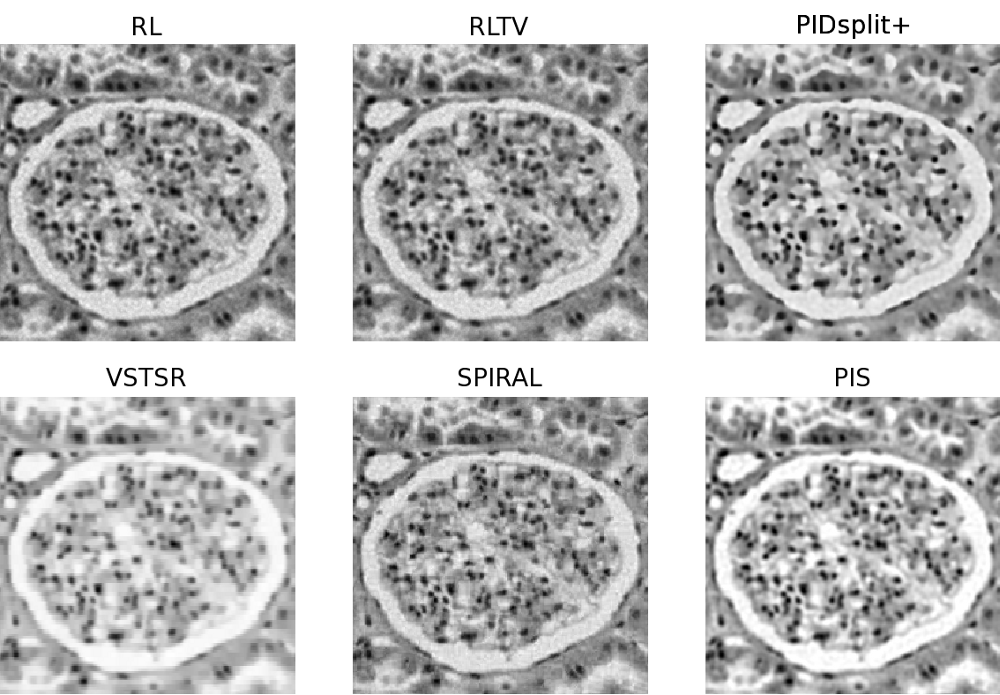} 
\caption{Image reconstruction results corresponding to Fig.\ref{F4} with D=2 and SNR=32. (Upper row of subplots) RL, RLTV and PIDsplit+ estimates; (Lower row of subplots) VSTSR, SPIRAL and PIS estimates.}
\label{F6}
\end{figure}

\begin{figure}[here]
\centering
\includegraphics[width=5in]{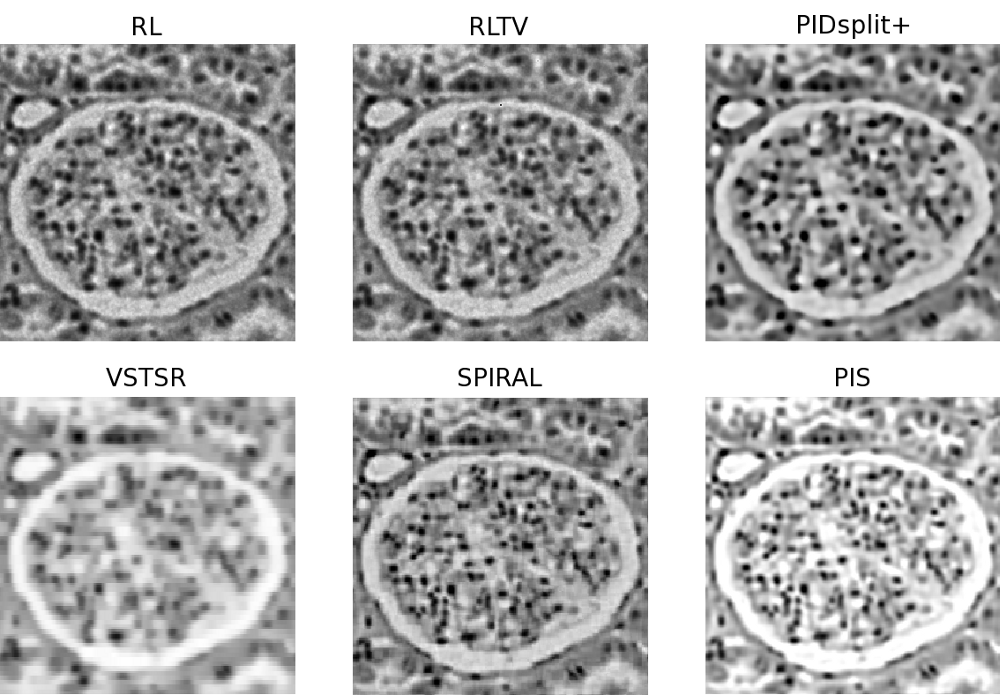}
\caption{Image reconstruction results corresponding to Fig.\ref{F4} with D=7 and SNR=8. (Upper row of subplots) RL, RLTV and PIDsplit+ estimates; (Lower row of subplots) VSTSR, SPIRAL and PIS estimates.}
\label{F7}
\end{figure}

\begin{figure}[htp]
\centering
\includegraphics[width=5in]{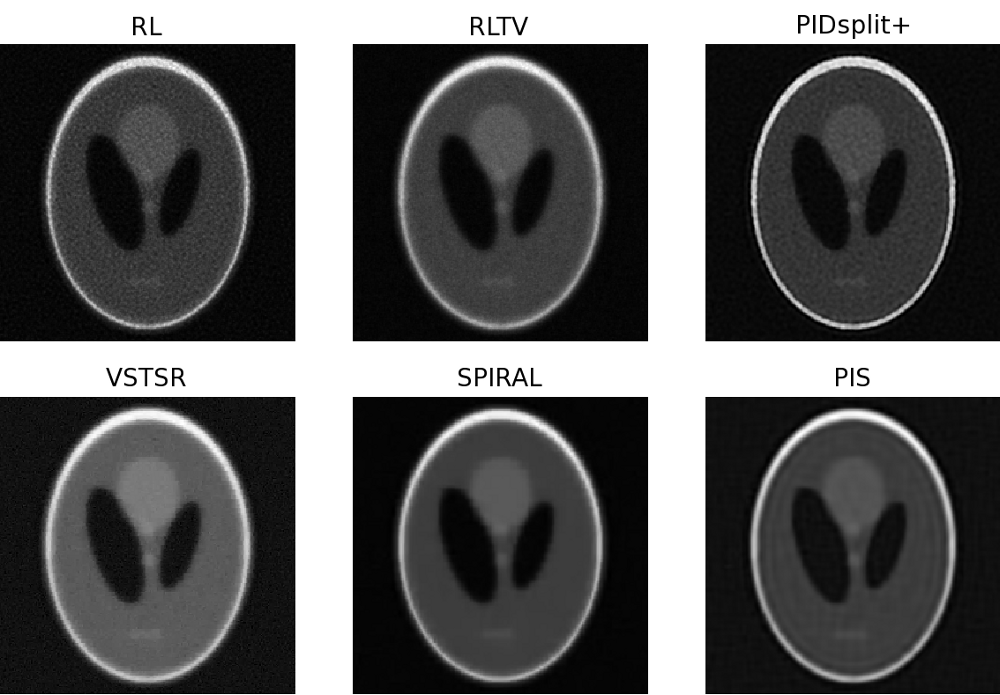} 
\caption{Image reconstruction results corresponding to Fig.\ref{F5} with D=2 and SNR=32. (Upper row of subplots) RL, RLTV and PIDsplit+ estimates; (Lower row of subplots) VSTSR, SPIRAL and PIS estimates.}
\label{F8}
\end{figure}

\begin{figure}[htp]
\centering
\includegraphics[width=5in]{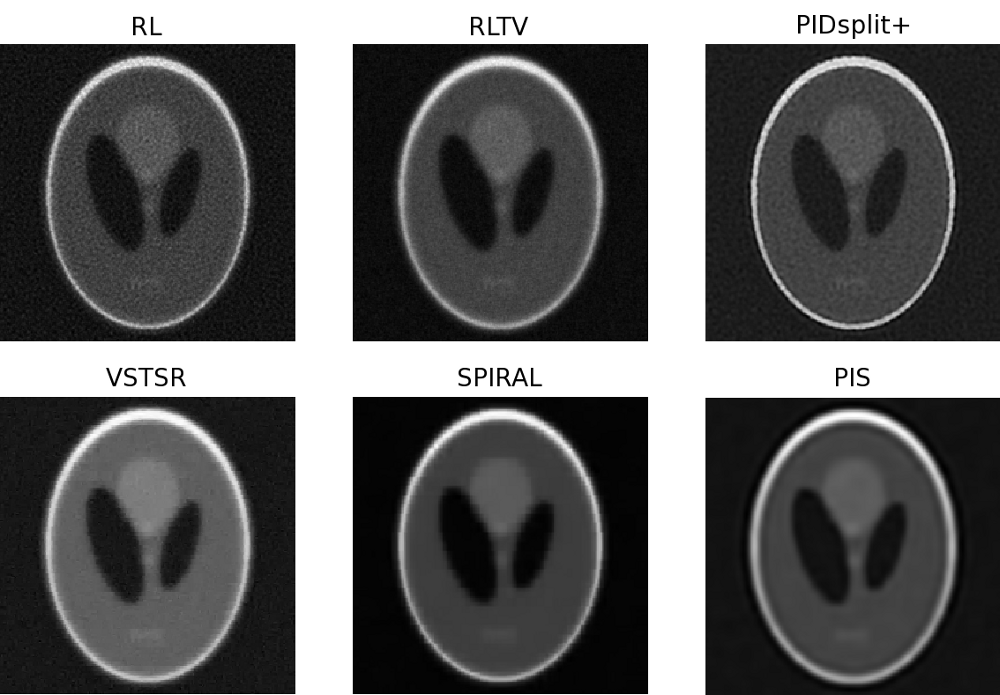}
\caption{Image reconstruction results corresponding to Fig.\ref{F5} with D=7 and SNR=8. (Upper row of subplots) RL, RLTV and PIDsplit+ estimates; (Lower row of subplots) VSTSR, SPIRAL and PIS estimates.}
\label{F9}
\end{figure}

The reference methods used in this section were RL, RLTV, PIDsplit+, VSTSR and SPIRAL. It should be noted that the GIS method has been excluded from the current experiment, whose inappropriateness of statistical model makes it a poor candidate for comparison (as demonstrated by the results of the previous section). The regularization parameters of VSTSR, SPIRAL and PIS were set empirically to be equal to 0.05, 0.1 and 0.02, respectively. The regularization parameters of RLTV and PIDsplit+ were set to be equal to 0.002 and 0.01 according the guidelines provided in \cite{Dey06} and \cite{Setzer09}.

The SPIRAL algorithm was applied with the orthogonal Haar wavelet transform (as opposed to its stationary version used by PIS), according to the requirements specified in \cite{Harmany09}. In this case, to alleviate the artifacts caused by the property of the orthogonal wavelet transform being translational  variant, the cycle-spinning algorithm of \cite{Antoniadis95} was employed, with a total number of cycles set to be equal to 20.  

In the case of the VSTSR, SPIRAL and PIS algorithms, their execution was terminated automatically at the point when the relative change $\| f_{t+1} - f_{t} \|_F \slash \| f_{t}\|_F $ between iterations $t$ and $t+1$ was observed to drop below a threshold of $10^{-6}$. Unfortunately, the same stopping criterion could not be applied to the RL, RLTV, and PIDsplit+ methods, whose steady-state estimation was found to be unacceptably noisy. For this reason, these algorithms were terminated earlier, at the point when their corresponding NMSE reached their minimum values. It should be noted that, since the computation of NMSE requires the knowledge of a true image, the ``NMSE-optimal" convergence cannot be considered as a practical tool. Therefore, the results of RL, RLTV, and PIDsplit+ methods reported in this section may not be reproduced in a real-life scenario.

For the case of glomerulus, the reconstructions obtained with the proposed and reference methods are summarized in Fig.~\ref{F6} (for $D=2$, SNR=32) and Fig.\ref{F7} (for $D=7$, SNR=8). Moreover, Fig.~\ref{F8} and Fig.~\ref{F9} depict the reconstructions of the Shepp-Logan phantom for the cases of $D=2$, SNR=32 and $D=7$, SNR=8, respectively. Analyzing these results, one can clearly see that, in all the above cases, the PIS algorithm yields reconstructions of superior quality (in terms of the resolution and contrast gain), as compared to the reference methods. This observation is further supported by the quantitative measures of Tables~\ref{T2} and~\ref{T3}, which compare the estimation results in terms of the NMSE, SSIM index, and the number of iterations. As evidenced by the tables, the PIS method produces the lowest NMSE and the largest SSIM index among all the methods under comparison. As to the number of iterations required by PIS, one can see (with a reference to Table~\ref{T1a}) that the method has a computational complexity either comparable or lower than that of the reference methods. 

\section{Discussion and Conclusions}
In the current paper, a new approach to the problem of denoising/reconstruction of digital images was presented. The method has been derived based on the framework of MAP estimation under the assumption of Poisson noise contamination. Such noise models are known to be standard in many important image modalities, including optical, microscopic, turbulent, and nuclear imaging, just to name a few. Moreover, whilst many of the existing solutions to the problem of enhancement of Poissonian images take advantage of certain simplifying assumptions about the noise nature, the proposed technique is optimized to deal with the realistic noise model at hand. 

Another advantage of the proposed method consists in the generality of its formulation. The latter allows applying the same reconstruction procedure to a number of different settings, such as image denoising or image enhancement through deconvolution. Furthermore, the prior assumptions made by the method regarding the nature of recovered images are general as well. Specifically, the images are assumed to admit a sparse representation in the domain of a properly chosen linear transform. Note that the reasonability of the above {\it a priori} modeling is firmly supported by the recent advances in the theory of sparse representation. 

Yet another critical advantage of the proposed PIS algorithm is in its algorithmic structure, which exploits the idea of iterative shrinkage. The latter allows solving non-smooth optimization problems at the computational cost of a steepest descent procedure. Consequently, the computational load required by the proposed method is relatively small, which allows the method to be applied for the solution of large-scale problems and/or for processing of large data sets. 

It was shown conceptually and experimentally that the performance of the PIS algorithm is superior to a number of alternative approaches. A series of comparison tests have been performed, in which PIS was shown to outperform the reference methods in terms of both NMSE and SSIM index measures. Moreover, as opposed to the alternative methods, the PIS algorithm has always been capable of converging in a stable and robust manner to a useful reconstruction result.  

We believe that the convergence speed of PIS iterations can be further improved via employing the line search strategy as detailed in~\cite{Elad06}. It seems also to be possible to increase the accuracy of PIS estimation by using more sparsifying multiresolution transforms as compared to separable wavelets~\cite{Candes02}. Finally, we note that the sparseness of representation coefficients appears to be a rather weak constraint to be used in the case of poorly conditioned convolution operators $\Hh$. This problem, therefore, should be resolved by using different prior models, whose formulation constitutes one of the directions of our current research.

\bibliographystyle{IEEEtran}

\begin{thebibliography}{10}
\providecommand{\url}[1]{#1}
\csname url@samestyle\endcsname
\providecommand{\newblock}{\relax}
\providecommand{\bibinfo}[2]{#2}
\providecommand{\BIBentrySTDinterwordspacing}{\spaceskip=0pt\relax}
\providecommand{\BIBentryALTinterwordstretchfactor}{4}
\providecommand{\BIBentryALTinterwordspacing}{\spaceskip=\fontdimen2\font plus
\BIBentryALTinterwordstretchfactor\fontdimen3\font minus
  \fontdimen4\font\relax}
\providecommand{\BIBforeignlanguage}[2]{{%
\expandafter\ifx\csname l@#1\endcsname\relax
\typeout{** WARNING: IEEEtran.bst: No hyphenation pattern has been}%
\typeout{** loaded for the language `#1'. Using the pattern for}%
\typeout{** the default language instead.}%
\else
\language=\csname l@#1\endcsname
\fi
#2}}
\providecommand{\BIBdecl}{\relax}
\BIBdecl

\bibitem{Lee95}
S.-J. Lee, A.~Rangarajan, and G.~Gindi, ``Bayesian image reconstruction in
  {SPECT} using higher order mechanical models as priors,'' \emph{{IEEE} Trans.
  Med. Imag.}, vol.~14, no.~4, pp. 669--680, Dec. 1995.

\bibitem{Yavuz98}
M.~Yavuz and J.~A. Fessler, ``Statistical image reconstruction methods for
  randomsprecorrected {PET} scans,'' \emph{Med. Im. Anal.}, vol.~2, no.~4, pp.
  369--378, 1998.

\bibitem{Bauschke99}
H.~H. Bauschke, D.~Noll, A.~Celler, and J.~M. Borwein, ``An {EM} algorithm for
  dynamic {SPECT},'' \emph{{IEEE} Trans. Med. Imag.}, vol.~18, no.~3, pp.
  252--261, Mar. 1999.

\bibitem{Boie92}
R.~A. Boie and I.~J. Cox, ``An analysis of camera noise,'' \emph{{IEEE} Trans.
  Pattern Anal. Machine Intell.}, vol.~14, no.~6, pp. 671--674, Jun. 1992.

\bibitem{Healey94}
G.~E. Healey and R.~Kondepudy, ``Radiometric {CCD} camera calibration and noise
  estimation,'' \emph{{IEEE} Trans. Pattern Anal. Machine Intell.}, vol.~16,
  no.~3, pp. 267--276, Mar. 1994.

\bibitem{Holmes92}
T.~J. Holmes, ``Blind deconvolution of quantum-limited incoherent imagery:
  {M}aximum-likelihood approach,'' \emph{J. Opt. Soc. Am. A}, vol.~9, no.~7,
  pp. 1052--1061, Jul. 1992.

\bibitem{Bradt04}
H.~Bradt, \emph{Astronomy methods: {A} physical approach to astronomical
  observations}.\hskip 1em plus 0.5em minus 0.4em\relax Cambridge University
  Press, 2004.

\bibitem{Roggeman96}
M.~C. Roggeman and B.~Welsh, \emph{Imaging through turbulence}.\hskip 1em plus
  0.5em minus 0.4em\relax CRC Press, 1996.

\bibitem{Bertero98}
M.~Bertero and P.~Boccacci, \emph{Introduction to inverse problems in
  imaging}.\hskip 1em plus 0.5em minus 0.4em\relax CRC Press, 1998.

\bibitem{Richardson72}
W.~H. Richardson, ``Bayesian-based iterative method of image restoration,''
  \emph{J. Opt. Soc. Am. A}, vol.~62, no.~1, pp. 55--59, 1972.

\bibitem{Lucy74}
L.~B. Lucy, ``An iterative technique for the rectification of observed
  distributions,'' \emph{Astron. J.}, vol.~79, no.~6, pp. 745--754, 1974.

\bibitem{Shepp82}
L.~A. Shepp and Y.~Vardi, ``Maximum likelihood reconstruction for emission
  tomography,'' \emph{{IEEE} Trans. Med. Imag.}, vol.~1, no.~2, pp. 113--122,
  Oct. 1982.

\bibitem{Mignotte02}
M.~Mignotte, J.~Meunier, J.-P. Soucy, and C.~Janicki, ``Comparison of
  deconvolution techniques using a distribution mixture parameter estimation:
  {A}pplication in single photon emission computed tomography imagery,''
  \emph{J. Electron. Imaging}, vol.~11, no.~1, pp. 11--24, Jan. 2002.

\bibitem{Dey06}
N.~Dey, L.~Blanc-Feraud, C.~Zimmer, P.~Roux, Z.~Kam, J.-C. Olivo-Marin, and
  J.~Zerubia, ``Richardson-{L}ucy algorithm with total variation regularization
  for 3-{D} confocal microscope deconvolution,'' \emph{Miscosc. Res. Techniq.},
  vol.~69, no.~4, pp. 260--266, Apr. 2006.

\bibitem{Figueiredo09}
M.~Figueiredo and J.~Bioucas-Dias, ``Deconvolution of {P}oissonian images using
  variable splitting and augmented {L}agrangian optimization,'' in \emph{IEEE
  Workshop on Statistical Signal Processing - SSP'2009}, Cardiff, Wales, UK,
  2009.

\bibitem{Goldstein09}
T.~Goldstein and S.~Osher, ``The split {B}regman method for {L}1 regularized
  problems,'' \emph{SIAM J. Imag. Sci.}, vol.~2, no.~2, pp. 323--343, 2009.

\bibitem{Setzer09}
S.~Setzer, G.~Steidl, and T.~Teuber, ``Deblurring {P}oissonian images by split
  {B}regman techniques,'' in \emph{Preprint, University of Mannheim}, 2009.

\bibitem{Lingenfelter09}
D.~Lingenfelter, J.~Fessler, and Z.~He, ``Sparsity regularization for image
  reconstruction with {P}oisson data,'' in \emph{Proceedings of SPIE
  (Computational Imaging VII)}, San Jose, California, USA, 2009.

\bibitem{Dupe08}
F.~X. Dupe, M.~J. Fadili, and J.~L. Strach, ``Image deconvolution under
  {P}oisson noise using sparse representations and proximal thresholding
  iteration,'' in \emph{ICASSP 2008}, Las-Vegas NV, USA, Apr. 2008.

\bibitem{Anscombe48}
F.~J. Anscombe, ``The transformation of {P}oisson, binomial, and negative
  binomial data,'' \emph{Biometrika}, vol.~35, pp. 246--254, 1948.

\bibitem{Harmany09}
Z.~Harmany, R.~Marcia, and R.~Willett, ``Sparse {P}oisson intensity
  reconstruction algorithms,'' in \emph{IEEE Stat. Sig. Proc. Workshop},
  Cardiff, Wales, UK, Aug. 2009.

\bibitem{TheWitch03}
R.~Willet and R.~D. Nowak, ``Platelets: {A} multiscale approach for recovering
  edges and surfaces in photon-limited medical imaging,'' \emph{{IEEE} Trans.
  Med. Imag.}, vol.~22, no.~3, pp. 332--350, 2003.

\bibitem{Antoniadis95}
R.~Coifman and D.~L. Donoho, \emph{Translation invariant denoising}, ser.
  Wavelets and statistics.\hskip 1em plus 0.5em minus 0.4em\relax New York,
  USA: Springer-Verlag, 1995.

\bibitem{Fryzlewicz04}
P.~Fryzlewicz and G.~P. Nason, ``A {H}aar-{F}isz algorithm for {P}oisson
  intensity estimation,'' \emph{J. Comput. Graph. Statist.}, vol.~13, pp.
  621--638, 2004.

\bibitem{Jansen06}
M.~Jansen, ``Multiscale {P}oisson data smoothing,'' \emph{J. Roy. Statist. Soc.
  B}, vol.~68, no.~1, pp. 27--48, 2006.

\bibitem{Fryzlewicz07}
P.~Fryzlewicz, V.~Delouille, and G.~P. Nason, ``{GOES}-8 {X}-ray sensor
  variance stabilization using the multiscale data-driven {H}aar-{F}isz
  transform,'' \emph{J. Roy. Statist. Soc. C}, vol.~56, no.~1, pp. 99--116,
  2007.

\bibitem{Zhang08}
B.~Zhang, J.~M. Fadili, and J.-L. Starck, ``Wavelets, ridgelets, curvelets for
  {P}oisson,'' \emph{{IEEE} Trans. Image Processing}, vol.~17, no.~7, pp.
  1093--1108, Jul. 2008.

\bibitem{Nowak99}
R.~D. Nowak and R.~G. Baraniuk, ``Wavelet-domain filtering for photon imaging
  systems,'' \emph{{IEEE} Trans. Image Processing}, vol.~8, no.~5, pp.
  666--678, May 1999.

\bibitem{Kolaczyk00}
E.~D. Kolaczyk, ``Nonparametric estimation of intensity maps using {H}aar
  wavelets and {P}oisson noise characteristics,'' \emph{Astrophys. J.}, vol.
  534, pp. 490--505, 2000.

\bibitem{Timmermann99}
K.~E. Timmermann and R.~D. Nowak, ``Multiscale modeling and estimation of
  {P}oisson processes with application to photo-limited imaging,'' \emph{{IEEE}
  Trans. Inform. Theory}, vol.~45, no.~3, pp. 846--862, Apr. 1999.

\bibitem{Nowak00}
R.~D. Nowak and E.~D. Kolaczyk, ``A statistical multiscale framework for
  {P}oisson inverse problems,'' \emph{{IEEE} Trans. Inform. Theory}, vol.~46,
  no.~5, pp. 1811--1825, Aug. 2000.

\bibitem{Elad07}
M.~Elad, B.~Matalon, J.~Shtok, and M.~Zibulevsky, ``A wide-angle view at
  iterated shrinkage algorithms,'' in \emph{Proceedings of SPIE (Wavelet XII)},
  San-Diego CA, USA, 2007.

\bibitem{Daubechies03}
I.~Daubechies, M.~Defrise, and C.~DeMol, ``An iterative thresholding algorithm
  for linear inverse problems with a sparsity constraint,''
  \emph{arXiv:math/0307152v2}, 2003.

\bibitem{Blumensath08}
T.~Blumensath and M.~E. Davies, ``Iterative thresholding for sparse
  approximations,'' \emph{J. Fourier Anal. Appl.}, vol.~14, no. 5-6, pp.
  629--654, Dec. 2008.

\bibitem{Wang04}
Z.~Wang, A.~C. Bovik, H.~R. Sheikh, and E.~P. Simoncelli, ``Image quality
  assessment: {F}rom error visibility to structural similarity,'' \emph{{IEEE}
  Trans. Image Processing}, vol.~13, no.~4, pp. 600--612, Apr. 2004.

\bibitem{Christensen08}
O.~Christensen, \emph{Frames and bases: {A}n introductory course}.\hskip 1em
  plus 0.5em minus 0.4em\relax Birkhauser, 2008.

\bibitem{Chen01}
S.~S. Chen, D.~L. Donoho, and M.~A. Saunders, ``Atomic decomposition by {B}asis
  {P}ursuit,'' \emph{SIAM Review}, vol.~43, no.~1, pp. 129--159, Mar. 2001.

\bibitem{Donoho06}
D.~Donoho, ``Compressed sensing,'' \emph{{IEEE} Trans. Inform. Theory},
  vol.~52, no.~4, pp. 1289--1306, Apr. 2006.

\bibitem{Candes06}
E.~Candes and T.~Tao, ``Near optimal signal recovery from random projections:
  {U}niversal encoding strategies,'' \emph{{IEEE} Trans. Inform. Theory},
  vol.~52, no.~12, pp. 5406--5425, Dec. 2006.

\bibitem{Bofill01}
P.~Bofill and M.~Zibulevsky, ``Underdetermined blind source separation using
  sparse representations,'' \emph{Signal Processing}, vol.~81, no.~11, pp.
  2353--2362, Nov. 2001.

\bibitem{Georgiev05}
P.~Georgiev, F.~Theis, and A.~Cichocki, ``Sparse component analysis and blind
  source separation of underdetermined mixtures,'' \emph{{IEEE} Trans. Neural
  Networks}, vol.~16, no.~4, pp. 992--996, Jul. 2005.

\bibitem{Figueiredo03}
M.~A.~T. Figueiredo and R.~D. Nowak, ``An {EM} algorithm for wavelet-based
  image restoration,'' \emph{{IEEE} Trans. Med. Imag.}, vol.~12, no.~8, pp.
  906--916, Aug. 2003.

\bibitem{Donoho95}
D.~L. Donoho, ``De-noising by soft-thresholding,'' \emph{{IEEE} Trans. Inform.
  Theory}, vol.~41, no.~3, pp. 613--627, May 1995.

\bibitem{Aharon06}
M.~Elad and M.~Aharon, ``Image denoising via sparse and redundant
  representations over learned dictionaries,'' \emph{{IEEE} Trans. Image
  Processing}, vol.~15, no.~12, pp. 3736--3745, Dec. 2006.

\bibitem{Starck04}
J.~L. Starck, M.~Elad, and D.~Donoho, ``Redundant multiscale transforms and
  their application for morphological component analysis,'' \emph{Adv. Imag.
  Electron. Phys.}, vol. 132, 2004.

\bibitem{Michailovich02}
O.~Michailovich and D.~Adam, ``A high-resolution technique for ultrasound
  harmonic imaging using sparse representations in {G}abor frames,''
  \emph{{IEEE} Trans. Med. Imag.}, vol.~21, no.~12, pp. 1490--1503, Dec. 2002.

\bibitem{Carlin96}
B.~P. Carlin and T.~A. Louis, \emph{Bayes and empirical {B}ayes methods for
  data analysis}, ser. Monographs on Statistics and Applied Probability.\hskip
  1em plus 0.5em minus 0.4em\relax Chapman and Hall, 1996, vol.~69.

\bibitem{Moulin99}
P.~Moulin and J.~Liu, ``Analysis of multiresolution image denoising schemes
  using generalized {G}aussian and complexity priors,'' \emph{{IEEE} Trans.
  Inform. Theory}, vol.~45, no.~3, pp. 909--919, Apr. 1999.

\bibitem{Boyd04}
S.~Boyd and L.~Vandenberghe, \emph{Convex optimization}.\hskip 1em plus 0.5em
  minus 0.4em\relax Cambridge University Press, 2004.

\bibitem{Bertsekas99}
D.~Bertsekas, \emph{Nonlinear programming}.\hskip 1em plus 0.5em minus
  0.4em\relax Belmont, MA: Athena Scientific, 1999.

\bibitem{Hunter04}
D.~R. Hunter and K.~Lange, ``A tutorial on {MM} algorithms,'' \emph{The
  American Statistician}, vol.~58, no.~1, pp. 30--37, Feb. 2004.

\bibitem{Figueiredo05}
M.~A.~T. Figueiredo and R.~D. Nowak, ``A bound optimization approach to
  wavelet-based image deconvolution,'' in \emph{Proceedings of ICIP}, Genova,
  Italy, 2005.

\bibitem{Gonzales02}
R.~C. Gonzales and R.~E. Woods, \emph{Digital image processing}.\hskip 1em plus
  0.5em minus 0.4em\relax Prentice hall, 2002.

\bibitem{Mallat99}
S.~Mallat, \emph{A wavelet tour of signal processing}.\hskip 1em plus 0.5em
  minus 0.4em\relax Academic Press, 1999.

\bibitem{Candes99}
E.~J. Candes and D.~L. Donoho, ``Ridgelets: {T}he key to high-dimensional
  intermittency?'' \emph{Philos. Trans. R. Soc. London, Ser. A}, vol. 357, pp.
  2495--2509, 1999.

\bibitem{Elad06}
M.~Elad, ``Why simple shrinkage is still relevant for redundant
  representations?'' \emph{{IEEE} Trans. Inform. Theory}, vol.~52, no.~12, pp.
  5559--5569, Dec. 2006.

\bibitem{Candes02}
E.~J. Candes and F.~Guo, ``New multiscale transforms, minimum total variation
  synthesis: {A}pplications to edge-preserving image reconstruction,''
  \emph{Signal Processing}, vol.~82, pp. 1519--1543, Mar. 2002.

\end{thebibliography}

\end{document}